\documentclass[letterpaper]{article} 
\pdfoutput=1
\usepackage{aaai2026}
\usepackage{times}  
\usepackage{helvet}  
\usepackage{courier}  
\usepackage[hyphens]{url}  
\usepackage{graphicx} 
\usepackage{natbib}  
\usepackage{caption} 
\usepackage[utf8]{inputenc}
\usepackage{graphicx}
\usepackage{amsmath}
\usepackage{cite}
\usepackage{booktabs}
\usepackage{array} 
\usepackage{algorithm}
\usepackage{algorithmic}

\usepackage{newfloat}
\usepackage{listings}
\DeclareCaptionStyle{ruled}{labelfont=normalfont,labelsep=colon,strut=off} 
\floatstyle{ruled}
\newfloat{listing}{tb}{lst}{}
\floatname{listing}{Listing}
\nocopyright

\title{Navigating the Trade-off: A Synthesis of Defensive Strategies for Zero-Shot Adversarial Robustness in Vision-Language Models}
\author{
    Zane Xu\textsuperscript{\rm 1}, Jason Sun\textsuperscript{\rm 2}\thanks{Corresponding author (1738400@park.edu).}
}
\affiliations{
    \textsuperscript{\rm 1}San Francisco State University\\
    \textsuperscript{\rm 2}Park University\\

}

\begin{document}

\maketitle


\begin{abstract}
This report synthesizes eight seminal papers on the zero-shot adversarial robustness of vision-language models (VLMs) like CLIP. A central challenge in this domain is the inherent trade-off between enhancing adversarial robustness and preserving the model's zero-shot generalization capabilities. We analyze two primary defense paradigms that have emerged from recent research. The first, Adversarial Fine-Tuning (AFT), involves modifying model parameters. We trace its evolution from early methods focused on protecting vision-language alignment (TeCoA) and mitigating overfitting through pre-trained model guidance (PMG-AFT, TGA-ZSR), to more advanced strategies that proactively re-engineer the embedding space's geometry to build intrinsic robustness (LAAT, TIMA). The second paradigm, Training-Free and Test-Time Defenses, avoids parameter modification altogether. We examine its progression from simple, heuristic-based input manipulations (AOM, TTC) to theoretically grounded purification methods in the model's latent space (CLIPure). By analyzing these works, we distill the field's core problems—including the robustness-generalization dilemma, geometric vulnerabilities, and computational costs—and identify key insights. We conclude by outlining promising future directions, such as the development of hybrid defense models and the pursuit of large-scale adversarial pre-training to create natively robust foundation models.
\end{abstract}

\section{Introduction}

\subsection{The Promise and Peril of Foundation Models}

In recent years, the field of artificial intelligence has witnessed a paradigm shift led by large-scale foundation models. Among these, Vision-Language Models (VLMs) such as CLIP (Contrastive Language-Image Pre-training)~\cite{CLIP} have been particularly noteworthy \cite{mao2023tecoa, wang2024pmg, li2024laat}. By performing contrastive learning on hundreds of millions of image-text pairs, models like CLIP have acquired unprecedented zero-shot generalization capabilities. They can achieve high-accuracy image classification on unseen downstream tasks and datasets using only simple text descriptions (e.g., "a photo of a dog"), significantly broadening the application boundaries of computer vision \cite{mao2023tecoa, tong2024aom}.

However, this powerful generalization capability conceals a major security vulnerability: extreme susceptibility to adversarial perturbations. Numerous studies have shown that adding tiny, human-imperceptible noise to an input image can cause the classification performance of these advanced models to drop precipitously, or even fail completely \cite{mao2023tecoa, ma2024tima}. For instance, in the CLIP model, a minute vector with a perturbation magnitude of no more than $1/255$ under the $l_{\infty}$ norm is sufficient to cause a subversive error in its prediction \cite{mao2023tecoa}. This vulnerability is not merely an academic issue; it poses a severe challenge to practical applications in safety-critical domains such as autonomous driving, medical diagnosis, and content moderation. Therefore, building robust defense mechanisms for these powerful foundation models has become an urgent research imperative.

\subsection{Defining the Core Challenge: The Robustness-Generalization Dilemma}

The concept of "zero-shot adversarial robustness" has emerged to address this need. Its core objective is to enhance a model's ability to withstand adversarial attacks while maximally preserving its invaluable zero-shot generalization capability. This goal is far more challenging than adversarial defense in traditional supervised learning scenarios because it introduces a fundamental dilemma: the inherent conflict between robustness and generalization.

Research has found that directly applying traditional adversarial defense methods, particularly Adversarial Training (AT), to foundation models like CLIP often leads to catastrophic consequences. For example, in a typical scenario, researchers conducted standard adversarial fine-tuning on the CLIP model using the ImageNet dataset. The results showed that while the model's adversarial robustness on the ImageNet validation set improved, its classification accuracy on 15 other unseen zero-shot datasets plummeted, with its generalization ability almost completely destroyed \cite{mao2023tecoa, wang2024pmg}. This phenomenon is known as catastrophic overfitting, indicating that the robust features learned through standard AT are highly coupled with the specific training data distribution, thereby corrupting the transferable knowledge gained from large-scale pre-training. How to resolve this "robustness-generalization" trade-off is the central challenge that the eight papers synthesized in this report collectively aim to address.

\subsection{A Taxonomy of Defense Paradigms}

To systematically organize and understand the various strategies for tackling this challenge, this report categorizes these eight pioneering works into two major defense paradigms. This classification is based on whether they require modification of the model's parameters to enhance robustness and at which stage of the model's lifecycle the defense mechanism operates.

\begin{itemize}
    \item \textbf{Paradigm I: Adversarial Fine-Tuning (AFT)}: The core idea of these methods is to "inject" robustness into the model itself by updating its parameters (i.e., fine-tuning) on a dataset containing adversarial examples. This process typically involves meticulously designed loss functions and regularization terms in the hope of learning robust features while avoiding catastrophic forgetting of pre-trained knowledge. The methods \textbf{TeCoA} \cite{mao2023tecoa}, \textbf{PMG-AFT} \cite{wang2024pmg}, \textbf{LAAT} \cite{li2024laat}, \textbf{TIMA} \cite{ma2024tima}, and \textbf{TGA-ZSR} \cite{yu2024tga} analyzed in this report belong to this paradigm.

    \item \textbf{Paradigm II: Training-Free and Test-Time Defenses}: These methods take a different path, completely avoiding any modification of the model's parameters. They intervene during the model's inference (test) stage by processing the input data or its internal representations to "purify" or "correct" the effects of adversarial perturbations. The advantage of this approach lies in its flexibility and low setup cost, as it bypasses the resource-intensive process of adversarial training. The methods \textbf{AOM} \cite{tong2024aom}, \textbf{TTC} \cite{xing2025ttc}, and \textbf{CLIPure} \cite{zhang2025clipure} analyzed herein are outstanding representatives of this paradigm.
\end{itemize}

The following sections will provide an in-depth analysis of the methods within these two paradigms and synthesize their findings to illuminate the core problems and future directions of the field.

\section{Paradigm I: Adversarial Fine-Tuning for Zero-Shot Robustness}

Adversarial Fine-Tuning (AFT) is a mainstream approach for enhancing model robustness. However, in the zero-shot setting, the challenge lies in guiding the model to learn generalizable robustness rather than merely memorizing attack patterns from a specific dataset. This section delves into five advanced AFT methods, revealing a progressive deepening of thought in solving the robustness-generalization dilemma.

\subsection{The Foundation: Preserving Vision-Language Alignment with TeCoA}

TeCoA \cite{mao2023tecoa} is one of the pioneering works in this field. It was the first to keenly observe that the reason standard adversarial training destroys CLIP's zero-shot capability is that it severs the core principle established during pre-training: vision-language alignment. Traditional adversarial training uses one-hot labels and cross-entropy loss, which decouples the optimization objective of the visual encoder from the text encoder, leading to the collapse of the joint embedding space.

\subsubsection*{Methodology}
To address this issue, TeCoA proposed a novel training objective: the \textbf{Text-guided Contrastive Adversarial (TeCoA) training loss}. Its core innovation is to shift the goal of adversarial training from "correctly classifying adversarial examples" to "maintaining correct vision-language alignment under adversarial perturbation." Specifically, this loss function is a cross-modal contrastive loss that aims to minimize the feature distance between an \textbf{adversarial image embedding} and its corresponding \textbf{correct text embedding}, while maximizing its distance from incorrect text embeddings. 

\subsubsection*{Significance}
TeCoA's contribution is foundational. It established a core principle: any adversarial fine-tuning method for vision-language models must make \textbf{preserving the joint embedding space} its central design goal. By reintroducing textual information into the adversarial training loop, TeCoA pointed the way for subsequent research.

\subsubsection*{Limitations}
Although TeCoA achieved significant improvements in zero-shot robustness over standard adversarial training, it still has clear limitations. Experiments show that models fine-tuned with TeCoA suffer a notable drop in clean accuracy compared to the original CLIP model and exhibit limited robustness against larger perturbations, indicating that overfitting remains an issue \cite{ma2024tima, wang2024pmg, li2024laat}. These shortcomings left room for improvement in subsequent research.

\subsection{Combating Overfitting: Guidance from the Pre-trained Model}

Building on TeCoA, subsequent researchers recognized that merely preserving the vision-language alignment objective was insufficient to completely combat overfitting. The reason is that the fine-tuning process can still cause the model parameters to "drift" too far from their pre-trained state. Therefore, a natural idea emerged: introduce direct supervision signals from the original pre-trained model during fine-tuning to constrain the learning process. PMG-AFT and TGA-ZSR are two different implementations of this idea, and together they reveal an evolutionary path of model regularization strategies from macroscopic to microscopic.

\subsubsection{Methodology (PMG-AFT)}
PMG-AFT (Pre-trained Model Guided Adversarial Fine-Tuning) \cite{wang2024pmg} proposes a direct and effective method to combat overfitting. It treats the original, frozen pre-trained CLIP model as a "teacher model" that continuously "guides" the "student model" being optimized during adversarial fine-tuning. This guidance is implemented through two core loss function components:
\begin{itemize}
    \item \textbf{Generalization Information Branch ($L_{general}$)}: This is the core innovation of PMG-AFT. It feeds adversarial examples into both the target model (student) and the original model (teacher) and then calculates the KL divergence between their output probability distributions. By minimizing this KL divergence, the loss function forces the student model's predictive behavior under adversarial attack to remain consistent with that of the more knowledgeable teacher model, thereby inheriting the teacher's generalization ability.
    \item \textbf{Regularization Loss ($L_{clean}$)}: This loss term encourages feature consistency within the target model. It calculates and minimizes the KL divergence between the target model's output probability distributions for the clean and adversarial versions of the same input image. This helps the model learn smooth feature representations that are insensitive to perturbations.
\end{itemize}

\subsubsection{Methodology (TGA-ZSR)}
TGA-ZSR (Text-Guided Attention for Zero-Shot Robustness) \cite{yu2024tga}, on the other hand, proposes a more refined form of supervision. Its core insight is that the success of an adversarial attack often manifests as an erroneous shift in the model's internal \textbf{text-guided attention map}. Therefore, instead of supervising the final output probabilities, it is more effective to directly supervise the model's internal "reasoning process." TGA-ZSR designs two modules to achieve this:
\begin{itemize}
    \item \textbf{Attention Refinement Module}: This module aims to correct the attention shifts caused by adversarial examples. It calculates the $L_2$ distance between the attention map generated by the target model on an \textbf{adversarial example} and the attention map generated by the original model on the corresponding \textbf{clean example}, and minimizes it as a loss term. This is equivalent to saying: "Even when under attack, you should focus on the correct regions of the image, just as the original model does on a clean image."
    \item \textbf{Attention-based Model Constraint Module}: To prevent the degradation of performance on clean samples while improving robustness, this module calculates and minimizes the $L_2$ distance between the attention maps generated by the target model and the original model on \textbf{clean examples}. This directly constrains the model's behavior when processing non-adversarial inputs, thereby preserving its generalization ability on clean data.
\end{itemize}

\subsubsection{The Evolution of Regularization from Macroscopic to Microscopic}
By analyzing TeCoA, PMG-AFT, and TGA-ZSR in sequence, we can clearly see an evolutionary path of regularization thinking from macroscopic to microscopic, from external to internal.
\begin{enumerate}
    \item The initial problem was that AFT destroyed CLIP's generalization ability.
    \item TeCoA's solution was to start at the \textbf{training objective} level, replacing cross-entropy with a contrastive loss to preserve CLIP's core property of vision-language alignment. This is a macroscopic, first-principles-based correction, but it lacks a direct constraint on the "drift" of model parameters, so the overfitting problem persisted.
    \item PMG-AFT noticed this and proposed a more direct regularization method: constraining the model's \textbf{final output}. It explicitly demands: "The final prediction of the fine-tuned model on an adversarial example should not deviate too far from the original model's prediction." This is a constraint on "what" the model does—its external behavior.
    \item TGA-ZSR goes a step further, hypothesizing that "why" a prediction is made is equally, if not more, fundamental. The constraint it proposes is: "The way the fine-tuned model \textit{looks at} an adversarial example (i.e., its attention) should not deviate too far from how the original model looks at a clean one." This is a constraint on "how" the model does it—its internal reasoning process.
\end{enumerate}
This shift from supervising "behavior" to supervising "thought" represents a deeper understanding of the zero-shot adversarial robustness problem. It suggests that to build a truly robust and generalizable model, one must not only teach it the right answers but also the right way to "think." This is an elevation from behavioral cloning to a more profound form of knowledge distillation, providing valuable insights for designing more powerful AFT methods.

\subsection{Reshaping the Embedding Space: Manipulating Text and Image Geometries}

The aforementioned methods primarily focus on "protecting" the knowledge learned by the pre-trained model during adversarial fine-tuning. However, another, more disruptive line of thought began to emerge: perhaps the embedding space learned by the pre-trained model, while optimized for zero-shot generalization, has fundamental geometric flaws when it comes to adversarial robustness. The works of LAAT and TIMA are representative of this idea; they are no longer content with passive defense and instead take the offensive, attempting to build intrinsic robustness by reshaping the geometry of the embedding space.

\subsubsection{Methodology (LAAT)}
LAAT (Language-driven, Anchor-based Adversarial Training) \cite{li2024laat} was the first work to diagnose a geometric flaw in CLIP's embedding space. It identified a key issue—the \textbf{"high cosine similarity problem."} Specifically, the text embeddings generated by CLIP for different classes are too clustered in the feature space, resulting in excessively high cosine similarity between them. This compact structure means that the decision boundaries between different classes are very narrow, providing convenience for the generation of adversarial examples.

To solve this problem, LAAT introduced a set of innovative solutions:
\begin{itemize}
    \item \textbf{Expansion Algorithm}: This is a novel geometric transformation algorithm designed to uniformly push the clustered text embeddings (which serve as "anchors" for classification) apart on the unit hypersphere, thereby increasing the inter-class distance and reducing cosine similarity. The algorithm performs a series of operations in spherical coordinates, such as rotation and enlargement of the polar angle, to expand the distance between anchors while striving to maintain their original relative topology to preserve semantic consistency.
    \item \textbf{Alignment Cross-Entropy (A-CE) Loss}: To better adapt to this expanded embedding space, LAAT employs an A-CE loss that combines cross-entropy and cosine similarity. Compared to a loss that directly maximizes the cosine similarity between the image feature and the corresponding text anchor, the A-CE loss provides a relaxation through the Softmax function, only requiring the similarity of the correct image-text pair to be higher than that of incorrect pairs, which offers greater flexibility for optimization.
\end{itemize}

\subsubsection{Methodology (TIMA)}
TIMA (Text-Image Mutual Awareness) \cite{ma2024tima} builds upon LAAT to construct a more comprehensive and systematic framework. It recognizes that the geometric vulnerability of the embedding space exists on both the \textbf{text side} and the \textbf{image side}, and proposes that they need to be co-optimized. TIMA consists of two mutually aware modules:
\begin{itemize}
    \item \textbf{Text-Aware Image (TAI) Tuning Module}: This module focuses on optimizing the decision boundaries in the image embedding space. Through empirical analysis, TIMA discovered two key phenomena: (1) there is a stable offset in the logit margin (the logit difference between the correct class and the most confusable class) when transitioning from weak to strong perturbations; (2) the logit margin is significantly negatively correlated with the semantic similarity between classes, meaning that semantically more similar classes have their margins more easily compressed by attacks. Based on these findings, the TAI module introduces the \textbf{Adaptive Semantic-Aware Margin (ASAM)}. ASAM enforces an adaptive, semantic-similarity-dependent negative margin in the contrastive loss, forcing the model to learn a larger safety gap for semantically closer classes, thereby directly reinforcing the decision boundaries on the image side.
    \item \textbf{Image-Aware Text (IAT) Tuning Module}: This module aims to address the issue that LAAT's expansion algorithm might disrupt the original semantic consistency. It proposes the \textbf{Semantic Consistent Minimum Hyperspherical Energy (SC-MHE)} method. This method includes two parts: one is the Minimum Hyperspherical Energy (MHE) term, which uses principles from physics to encourage a uniform distribution of text embeddings on the hypersphere, thereby maximizing inter-class distance; the other is a crucial \textbf{semantic consistency regularization term}, which explicitly preserves the valuable semantic structure learned by the pre-trained model by constraining the predictive distribution of the tuned text embeddings on clean images to be consistent with that of the original embeddings.
\end{itemize}

\subsubsection{The Leap in Thinking from Corrective Measures to Proactive Design}
Comparing methods like TeCoA and PMG-AFT with LAAT and TIMA, we can observe a major leap in research thinking: from taking \textbf{corrective measures} to conducting \textbf{proactive design}.
\begin{enumerate}
    \item Early AFT methods (like TeCoA, PMG-AFT) implicitly assumed that the CLIP pre-trained embedding space is "good," and their core goal was to "not damage it" during fine-tuning. Their strategies were passive and protective.
    \item LAAT challenged this assumption. By diagnosing the "high cosine similarity problem," it was the first to explicitly state that the pre-trained text embedding space is geometrically flawed for robustness. Its proposed expansion algorithm is a \textbf{corrective measure}—find a problem, then fix it.
    \item TIMA pushed this idea to maturity. It not only recognized that both text and image spaces have geometric vulnerabilities, but its proposed solutions are more akin to a \textbf{proactive system design}. The ASAM module does not wait for a problem to arise to fix it; it proactively designs more robust decision boundaries based on semantic relationships. The SC-MHE module, while pursuing the goal of increasing inter-class distance, has a built-in constraint to maintain semantic consistency, which is a geometric reconstruction with clear design objectives.
    \item This evolutionary trajectory clearly shows the maturation of the research thinking: from an initial "do no harm" philosophy to an "actively re-engineer for robustness" philosophy. The "Text-Image Mutual Awareness" framework proposed by TIMA, through the co-optimization of both modalities, represents the state of the art in this AFT paradigm and signifies a deepening of the understanding of the problem's root cause.
\end{enumerate}

\section{Paradigm II: Training-Free and Test-Time Defenses}

In contrast to the resource-intensive adversarial fine-tuning paradigm, training-free and test-time defenses offer a more lightweight and flexible path. The core advantage of these methods is that they do not alter any parameters of the pre-trained model, thus fundamentally avoiding catastrophic forgetting and overfitting issues. The defense mechanism is activated during the model's inference stage, mitigating the impact of adversarial perturbations by processing the input sample or its feature representation in real-time. This section will explore three representative methods—AOM, TTC, and CLIPure—and reveal their different levels of theoretical grounding and implementation complexity.

\subsection{Purification in Latent Space: A Shift from Pixels to Semantics}

Traditional adversarial purification methods mostly operate in pixel space, attempting to restore an adversarial image to its corresponding clean version. However, the high dimensionality and sparsity of pixel space make this task exceptionally difficult. CLIPure \cite{zhang2025clipure} proposes a paradigm-level shift: moving the battlefield of purification from the high-dimensional pixel space to the low-dimensional, dense, and more semantically meaningful CLIP latent space.

\subsubsection*{Methodology (CLIPure)}
The theoretical cornerstone of CLIPure lies in its mathematical modeling of the purification process using Stochastic Differential Equations (SDEs). It models the attack process (adding perturbations to clean samples) as a forward SDE and the purification process (denoising adversarial samples) as a reverse SDE. By analyzing the KL divergence between the joint distributions of these two processes, CLIPure derives that the lower bound of the purification risk is related to two key factors: (1) the difference in probability distributions between clean and adversarial samples; (2) the $l_2$ norm of the gradient of the adversarial sample's probability distribution. This theoretical analysis strongly supports its core argument: purification in a smoother, denser latent space carries a much lower risk than in pixel space.

Based on this theory, CLIPure proposes two methods for estimating the likelihood of "clean" image embeddings in the CLIP latent space, using this likelihood as the objective for gradient ascent to purify the adversarial embedding:
\begin{itemize}
    \item \textbf{CLIPure-Diff}: This is a \textbf{generative} approach. It cleverly utilizes the Diffusion Prior module from the DALL-E 2 model, which is itself a diffusion model trained on the CLIP latent space. CLIPure-Diff conditions on the text embedding of a null template (e.g., "a photo of a.") to estimate the log-likelihood of a given image embedding, thereby guiding the purification process.
    \item \textbf{CLIPure-Cos}: This is a highly innovative \textbf{discriminative} approach. It proposes a simple yet effective proxy for likelihood: the "cleanliness" of an image embedding can be measured by its \textbf{cosine similarity} to the text embedding of the aforementioned null template. This method completely eliminates the reliance on large and slow-to-infer generative models, improving the efficiency of purification by several orders of magnitude. Its inference time is only 1.14 times that of the original CLIP \cite{zhang2025clipure}.
\end{itemize}

\subsubsection*{Significance}
The work of CLIPure is doubly significant. First, it provides a solid theoretical basis for "why latent space defense is superior," formalizing an intuitive idea into a rigorous mathematical framework. Second, the proposal of CLIPure-Cos pioneers a new paradigm of adversarial purification that does not rely on generative models, greatly enhancing the practicality and feasibility of test-time defense.

\subsection{Input and Feature Manipulation at Inference Time}

Unlike CLIPure's probability-model-based purification, AOM and TTC adopt more direct, heuristic-based strategies for manipulating features or inputs. They design corresponding "countermeasures" by observing the specific behavior of adversarial examples at inference time.

\subsubsection{Methodology (AOM)}
The method of AOM (Anchor-guided One-step linear Movement) \cite{tong2024aom} originates from a simple yet effective empirical observation: adding a small amount of Gaussian noise to an adversarial example can, to some extent, weaken the effect of the adversarial perturbation and improve the model's recognition accuracy. Based on this, AOM designs a three-step defense process:
\begin{enumerate}
    \item \textbf{Establish the Source Point}: The input (potentially adversarial) image is treated as the source point.
    \item \textbf{Construct the Anchor}: To find a "cleaner" feature direction, AOM adds different Gaussian noises to the source image multiple times, feeds these noisy images into CLIP's visual encoder to obtain multiple feature embeddings, and then averages these embeddings to get an "anchor" feature. This anchor is considered to be in a more robust region near the source feature.
    \item \textbf{One-step Linear Movement}: Finally, in the feature space, AOM performs a single linear interpolation to move the source point's feature embedding along the direction pointing to the anchor feature. This new, moved feature point is then used as the final, defended feature for subsequent classification.
\end{enumerate}

\subsubsection{Methodology (TTC)}
TTC (Test-time Counterattacks) \cite{xing2025ttc} proposes a more adversarial defense philosophy. It is based on a more subtle observation, the phenomenon of \textbf{"false stability"}: adversarial examples generated by maximizing classification loss (like PGD) exhibit an unusual "stability" to small random noises in the latent space, meaning their feature embeddings drift to a much lesser extent than clean samples. TTC interprets this phenomenon as the adversarial example being "trapped" in a "toxic" local region created by the attacker.

To allow the sample to "escape" this region, TTC launches a "counterattack" at test time:
\begin{itemize}
    \item \textbf{Counterattack Mechanism}: Using the input image's own embedding as an anchor, TTC uses an optimization process (PGD) to find a "counterattack perturbation." The goal of this perturbation is to \textbf{maximize} the $L_2$ distance between the image embedding after adding the perturbation and the original anchor embedding. This is equivalent to applying a force at the input end to push the feature point in the latent space away from its current position.
    \item \textbf{$\tau$-threshold Mechanism}: To prevent this "counterattack" from harming innocent clean samples, TTC designs a clever switch. Before launching the counterattack, it first measures the "stability" of the input sample (the $\tau$ value) by adding a tiny random noise. The counterattack procedure is activated only when the stability is higher than a certain threshold (i.e., it exhibits "false stability"). This greatly protects the model's performance on clean samples.
\end{itemize}

\subsection{A Spectrum from Empirical Observation to Theoretical Formalism}
Analyzing AOM, TTC, and CLIPure side-by-side reveals a clear spectrum from simple empirical observation to complex theoretical formalism, reflecting a continuous deepening of researchers' understanding of the problem.
\begin{enumerate}
    \item The high cost and overfitting risks of adversarial fine-tuning created a demand for training-free alternatives.
    \item \textbf{AOM} provides the simplest proof-of-concept. It is based on a direct empirical finding—"Gaussian noise works"—and translates it into a simple feature space manipulation algorithm. The justification for its method is purely empirical.
    \item \textbf{TTC} proposes a more sophisticated hypothesis about the \textit{nature} of adversarial examples in CLIP's latent space—"false stability." Its defense is a direct countermeasure to this observed property. While still heuristic-driven, its motivation is deeper than AOM's, attempting to explain an intrinsic behavioral pattern of adversarial examples.
    \item \textbf{CLIPure} ultimately provides the underlying theory. Its SDE-based risk analysis mathematically explains \textit{why} the latent space is a better place to operate than the pixel space. It formalizes the common goal of all purification methods (maximizing the likelihood of clean samples) and provides two principled ways to achieve it. From this perspective, AOM and TTC can be seen as practical, efficient approximations of the general principle of purification that CLIPure formalizes.
\end{enumerate}

\section{Synthesis and Analysis of Core Challenges}

Through an in-depth analysis of the eight representative methods under the two major paradigms, we can form a comprehensive and multi-faceted understanding of the key challenges in the field of zero-shot adversarial robustness. These challenges are not isolated but intertwined, collectively forming the core driving force of research in this area.

\subsection{A Comparative Analysis of Methodologies}

To intuitively compare the similarities and differences of these eight methods, the following table summarizes their core characteristics across multiple dimensions. This table is not just a list of information but an analytical tool that reveals the focus and evolutionary trends of different technical routes.

\begin{table*}[!ht]
\centering
\tabcolsep=3.7pt
\extrarowheight=-0.8pt
\caption{Comparative Analysis of Zero-Shot Adversarial Defense Methodologies}
\label{tab:method_comparison}
\begin{tabular}{p{2.5cm} p{3.5cm} p{4cm} p{3cm} p{3.5cm}}
\toprule
\textbf{Method} & \textbf{Core Paradigm} & \textbf{Key Innovation} & \textbf{Modified Component} & \textbf{Primary Challenge Addressed} \\
\midrule
\textbf{TeCoA}   & Adversarial Fine-Tuning & Text-guided contrastive loss, preserving alignment & Image Encoder & Alignment Destruction \\
\textbf{PMG-AFT}   & Adversarial Fine-Tuning & Original model guidance, constraining output distribution & Image Encoder & Overfitting \\
\textbf{TGA-ZSR}  & Adversarial Fine-Tuning & Original model guidance, constraining attention mechanism & Image Encoder & Overfitting, Interpretability \\
\textbf{LAAT}   & Adversarial Fine-Tuning & Identifying and solving the "high cosine similarity problem" & Image Encoder, Text Embeddings & Embedding Space Geometry Flaws \\
\textbf{TIMA}   & Adversarial Fine-Tuning & Text-image mutual awareness, co-optimizing boundaries and distribution & Image Encoder, Text Embeddings & Embedding Space Geometry Flaws, Generalization-Robustness Trade-off \\
\textbf{AOM}   & Training-Free & Feature space movement based on Gaussian noise anchor & Input Features at Test-Time & Efficiency, Simplicity \\
\textbf{TTC}   & Training-Free & Test-time counterattack based on "false stability" & Input Image at Test-Time & Efficiency, Preserving Clean Performance \\
\textbf{CLIPure}   & Training-Free & Latent space purification, non-generative likelihood estimation & Input Features at Test-Time & Efficiency, Theoretical Foundation \\
\bottomrule
\end{tabular}
\end{table*}

\subsection{The Robustness-Generalization Dilemma Revisited}

All eight works attempt to resolve the fundamental conflict between robustness and generalization, but they adopt two distinct macro-strategies.

\begin{itemize}
    \item \textbf{The "Mitigate Damage" Strategy of Adversarial Fine-Tuning}: Methods under the AFT paradigm (TeCoA, PMG-AFT, LAAT, TIMA, TGA-ZSR) acknowledge that parameter updates are a necessary means to learn robustness, and thus they accept the potential loss of generalization ability that fine-tuning may bring. Their core task is to \textbf{maximally mitigate} this damage through various sophisticated regularization terms (like PMG-AFT's model guidance \cite{wang2024pmg}), geometric correction techniques (like LAAT's expansion algorithm \cite{li2024laat}), or co-optimization frameworks (like TIMA's mutual awareness \cite{ma2024tima}). Their success is measured by their ability to achieve higher robustness while retaining more zero-shot generalization capability compared to standard adversarial training.

    \item \textbf{The "Evade Trade-off" Strategy of Training-Free Methods}: Methods under the training-free paradigm (AOM, TTC, CLIPure) attempt to completely \textbf{evade} this trade-off at the training stage. By postposing the defense mechanism to the inference stage, they ensure that the pre-trained model's generalization ability remains intact. However, this is not without cost. They face a new, \textbf{inference-time trade-off}: the defense mechanism must be effective on adversarial inputs while absolutely not harming performance on clean inputs. The $\tau$-threshold mechanism proposed by TTC \cite{xing2025ttc} is a direct solution designed for this inference-time dilemma. It decides whether to activate the defense by identifying "false stability," thereby protecting clean samples from unnecessary processing.
\end{itemize}

\subsection{The Battleground of Embedding Spaces}

The combined contributions of LAAT, TIMA, and CLIPure profoundly reveal that the core battleground for zero-shot adversarial robustness lies in the intrinsic properties of the vision-language joint embedding space. Robustness is not just a matter of the classification loss function, but a matter of the geometric structure of the embedding space.

\begin{itemize}
    \item \textbf{Vulnerabilities on the Text Side}: LAAT \cite{li2024laat} and TIMA \cite{ma2024tima} jointly identified two major vulnerabilities on the text embedding side. One is a \textbf{geometric structure flaw}, the "high cosine similarity problem," which leads to insufficient discrimination between classes. The other is the \textbf{semantic consistency risk}, meaning that in correcting the geometric structure, one might unintentionally destroy the valuable semantic relationships learned by the pre-trained model (e.g., the spatial proximity of "cat" and "dog" embeddings). TIMA's SC-MHE module provides an effective solution to this risk by introducing an explicit regularization term.

    \item \textbf{Vulnerabilities on the Image Side}: TIMA \cite{ma2024tima} further revealed the vulnerability on the image embedding side, namely \textbf{insufficient logit margins}. Especially for semantically similar classes, their decision boundaries are too narrow and are easily crossed by adversarial attacks. TIMA's ASAM module directly enhances the defense on the image side by adaptively widening the decision boundaries in these critical regions.

    \item \textbf{The Superiority of the Latent Space}: CLIPure \cite{zhang2025clipure} then argued from a more macroscopic perspective that the entire CLIP latent space (both image and text) as a whole is a far superior defense battlefield than the pixel space. Its lower dimensionality, smoother manifold, and richer semantic structure make the task of distinguishing and purifying adversarial perturbations a more tractable problem.
\end{itemize}

\subsection{The Rise of Training-Free Defenses: A Paradigm Shift?}

The successive emergence of AOM, TTC, and CLIPure marks an important trend in the field. This trend is driven by the urgent need for more practical, efficient, and scalable defense solutions and may herald a paradigm shift.

\begin{itemize}
    \item \textbf{Advantages}: Training-free methods have significant advantages. First, they have \textbf{no catastrophic forgetting} problem because the model parameters remain unchanged. Second, their \textbf{deployment cost is lower}, as they do not require expensive adversarial fine-tuning and can be flexibly applied to any off-the-shelf pre-trained model. Finally, they have \textbf{better adaptability}, theoretically being plug-and-play for future new models.

    \item \textbf{Challenges}: However, this paradigm also faces its own severe challenges. The most significant is the \textbf{inference-time computational overhead}. Although methods like AOM and TTC are relatively lightweight, methods like CLIPure-Diff that rely on large generative models can be very time-consuming. The proposal of CLIPure-Cos \cite{zhang2025clipure} is precisely to address this efficiency bottleneck. A more fundamental challenge is that all test-time defense methods face the threat of \textbf{adaptive attacks}. An attacker who knows the specific details of the defense mechanism can design adversarial examples specifically to bypass it. Therefore, the robustness evaluation of these training-free methods must be conducted under strong adaptive attacks to verify their effectiveness in real-world security scenarios.
\end{itemize}

\section{Key Problems, Insights, and Future Directions}

After a systematic review and comprehensive analysis of the eight pioneering papers, we can distill the core problems, key insights, and envision future research directions for this field.

\subsection{Recapitulation of Key Problems}

These research works collectively focus on several fundamental challenges:
\begin{enumerate}
    \item \textbf{The Robustness-Generalization Trade-off}: This is the central conflict of the field. How to improve the model's ability to withstand adversarial attacks without sacrificing its zero-shot generalization ability on unknown tasks?
    \item \textbf{Overfitting}: Adversarial fine-tuning methods are naturally at risk of overfitting to the fine-tuning dataset, which directly leads to a decline in generalization ability.
    \item \textbf{Computational Cost}: The adversarial fine-tuning process is resource-intensive, while training-free methods shift this cost to every inference. Achieving a balance between effectiveness and efficiency is a key engineering problem.
    \item \textbf{Geometric Vulnerabilities}: The pre-trained embedding space is not inherently designed for adversarial robustness; its geometric structure (e.g., small inter-class distances, narrow decision margins) has intrinsic vulnerabilities.
    \item \textbf{Interpretability}: A deep understanding of why attacks succeed (e.g., the attention shift phenomenon discovered by TGA-ZSR \cite{yu2024tga}) can inspire the design of more targeted and effective defense strategies.
\end{enumerate}

\subsection{Core Insights Derived from the Literature}

Synthesizing the contributions of all the papers, we can draw the following profound insights:
\begin{itemize}
    \item \textbf{Preserving Pre-trained Knowledge is Paramount}: Successful adversarial fine-tuning is not about "re-learning," but about robustly adapting while heavily regularizing towards the knowledge of the original model. Whether constraining the output distribution (PMG-AFT \cite{wang2024pmg}) or the internal attention (TGA-ZSR \cite{yu2024tga}), the core is to learn from the pre-trained "teacher."
    \item \textbf{Geometry is as Important as Alignment}: A model's robustness depends not only on whether the vision and language features are aligned but also profoundly on the geometric structure of the embedding space—including the distances between class prototypes, the width of decision margins, and the distribution of features. Directly manipulating and optimizing these geometric properties is an extremely powerful defense strategy (LAAT \cite{li2024laat}, TIMA \cite{ma2024tima}).
    \item \textbf{Defense Can Be Training-Free}: Adversarial robustness is not exclusively attainable at the training stage. By purifying, correcting, or counteracting inputs/features at inference time, a practical and efficient alternative can be provided, effectively circumventing many of the drawbacks of adversarial fine-tuning (AOM \cite{tong2024aom}, TTC \cite{xing2025ttc}, CLIPure \cite{zhang2025clipure}).
    \item \textbf{Interpretability Drives Innovation}: A deep understanding of attack mechanisms is a source of innovation. By explaining how attacks affect the internal workings of a model, researchers can discover new attack surfaces and corresponding defense strategies, as inspired by the attention shifts in TGA-ZSR \cite{yu2024tga}.
\end{itemize}

\subsection{Future Research Horizons}

Based on the above analysis, future research could seek breakthroughs in the following directions:
\begin{itemize}
    \item \textbf{Hybrid Defense Models}: Combining the strengths of AFT and test-time defenses may be a fruitful direction. For example, one could first perform a lightweight, geometry-focused fine-tuning on a model using TIMA's principles \cite{ma2024tima} to build a more robust base model, and then deploy a fast test-time defense module like TTC \cite{xing2025ttc} or CLIPure-Cos \cite{zhang2025clipure} at inference time to form a dual-layer protection. TTC's own results on AFT models suggest this is a promising direction \cite{xing2025ttc}.
    \item \textbf{Developing and Evaluating Against Adaptive Attacks}: For training-free methods in particular, current research is mostly evaluated under standard, non-adaptive attacks. Future work must shift its focus to developing adaptive attack algorithms specifically designed to circumvent these defense mechanisms and conducting rigorous robustness evaluations under them. This is a critical next step to validate their effectiveness in real-world security confrontations.
    \item \textbf{Large-Scale Adversarial Pre-training}: All the papers reviewed focus on the "downstream" robustness modification of an existing pre-trained CLIP model. An ultimate, yet highly challenging, question is: can we incorporate adversarial robustness into the model during the pre-training stage itself? That is, performing large-scale adversarial pre-training from scratch on the 400-million-level image-text pair dataset. Although this faces enormous computational resource challenges and is noted as a limitation by some researchers \cite{ma2024tima}, it represents the "holy grail" of the field—building a foundation model that is robust by default and requires no additional defense.
    \item \textbf{Extending Beyond Classification}: The vast majority of current research focuses on zero-shot image classification tasks. An important future direction is how to generalize these principles of preserving alignment, correcting geometry, and test-time purification to other tasks that require zero-shot robustness, such as adversarial defense for object detection, segmentation, and even for text-to-image generation models.
\end{itemize}

\section{Conclusion}

This report has provided a systematic review and analysis of eight pioneering defense works on the zero-shot adversarial robustness of vision-language models. The research shows that two main defense paradigms have formed in this field: \textbf{Adversarial Fine-Tuning} and \textbf{Training-Free/Test-Time Defense}.

Within the \textbf{Adversarial Fine-Tuning} paradigm, the research thinking has undergone a profound evolution: from initially striving to \textbf{passively protect} the pre-trained vision-language alignment through special loss functions (TeCoA), to \textbf{actively regularizing} the learning process by introducing the original model as supervision (PMG-AFT, TGA-ZSR), and finally moving towards \textbf{proactively designing} intrinsic robustness by manipulating the geometric properties of the vision-language embedding space (LAAT, TIMA). This evolutionary path reflects a continuous deepening of the understanding of the problem's root cause, moving from focusing on "behavior" to "mechanism," and then to "structure."

Meanwhile, the rise of the \textbf{Training-Free and Test-Time Defense} paradigm offers a brand-new perspective to solve the inherent high cost and overfitting problems of AFT methods. Whether it is heuristic methods based on empirical observations (AOM, TTC) or latent space purification based on rigorous theory (CLIPure), they all demonstrate the great potential of achieving robustness without modifying model parameters, representing an important trend for future defense technologies to develop towards higher efficiency and flexibility.

In conclusion, zero-shot adversarial robustness is a challenging but crucial research area. Through a comprehensive analysis of existing work, we believe that the future development of this field likely lies in the construction of \textbf{hybrid models}, which combine the structural optimization of AFT with the flexibility of test-time defense. The ultimate ideal goal, however, is to build the next generation of foundation models that are natively robust from the source through \textbf{large-scale adversarial pre-training}. These explorations will continue to drive us to build safer, more reliable, and more trustworthy artificial intelligence systems.

\bibliography{aaai2026}



\end{document}